\documentclass{styles/svproc}
\usepackage{url}
\usepackage{tcolorbox}
\usepackage{graphicx}
\usepackage{subcaption}
\usepackage{booktabs}
\usepackage{multirow}

\begin{document}
\mainmatter              
%
\title{Exploring the link between predictive uncertainty and the sim-to-real gap in deep learning models utilizing synthetic data}
%


\titlerunning{Predictive uncertainty \& sim-to-real gap}

\author{Alexander Moriz$^{1}$, Lukas Krebs$^{1}$, Amon Göppert$^{1}$,
        Robert H. Schmitt$^{1,2}$\\[1.5ex]
  {\small
  $^{1}$Chair for Intelligence in Quality Sensing (IQS),
  WZL \textbar\ RWTH Aachen University,
  Campus Boulevard 30, 52074 Aachen, Germany\\
  $^{2}$Fraunhofer Institute for Production Technology IPT,
  Steinbachstrasse 17, 52074 Aachen, Germany\\[1ex]
  \texttt{alexander.moriz@wzl-iqs.rwth-aachen.de}}}

\authorrunning{Alexander Moriz et al.}

\maketitle             

\begin{abstract}
Today, state-of-the-art methods for quality control (QC) are based on Machine Vision, combining deep learning (DL) models with camera-based inspection. 
Yet, the integration of such solutions in industrial applications is often impeded by data sparsity and trust in model reliability. 
While synthetic data offers a promising approach to generate the necessary data with perfect annotations, it poses risks related to performance degradation due to the possible occurrence of the sim-to-real gap. 
This study conducts experiments using the ensemble method to estimate the predictive uncertainty of DL models trained on synthetic data. 
The results indicate a strong link between the sim-to-real gap and a decreased model performance, notably exceeding the estimated epistemic uncertainty.
These findings contribute to the effective utilization of synthetic data in industrial settings while highlighting the importance of risk assessment for the deployment of DL solutions for QC tasks.

\end{abstract}
\section{Introduction}

In recent years, the utilization of deep learning (DL) in various industrial sectors has gained significant momentum.
Especially for quality control (QC) applications, Machine Vision (MV) solutions, combining camera-based inspection and DL evaluation, are considered state-of-the-art methods \cite{kim:2025}.
However, the integration of such MV approaches is often impeded by data sparsity, a lack of expertise, and a lack of trust in the reliability of DL \cite{Ahangar:2025}.
In assembly environments, trust in the behavior of employed approaches is particularly crucial, as a failing QC system may allow incorrectly or incompletely assembled products to proceed downstream, potentially resulting in costly rework, product recalls, or safety hazards for end users \cite{Baltag:2022}.
One way to establish trust in DL models is through a thorough risk assessment, including, especially, an uncertainty estimation of the model's behavior.
Yet, acquiring such assessments for DL models remains a challenging and labor-intensive endeavor and is therefore often omitted both in the literature and praxis \cite{Gawlikowski:2022}.

Synthetic data presents a promising approach that addresses the data sparsity challenge, offering the generation of an arbitrary amount of data with perfect annotations for the training of DL models \cite{akar:2022}.  This is especially relevant for assembly QC, where the high variability of possible assembly errors, such as missing, misaligned, or incorrectly oriented components, makes comprehensive real-world data collection impractical. 
Furthermore, by utilizing synthetic data not only for training but also for evaluation, the controlled and known distribution properties of the synthetic data offer a unique opportunity to isolate and investigate the predictive uncertainty of DL models. 
The disadvantage of synthetic data as a training basis is the risk of performance degradation when applied to real-world data due to a potential sim-to-real gap between the real-world and the simulation.
In addition, the linkage between the inherent predictive uncertainty of synthetically trained DL models and the influence of a potential sim-to-real gap remains largely unexplored \cite{Thuy:2024}.

This work investigates this connection by conducting a series of experiments to estimate the predictive uncertainty of synthetically trained DL models using the ensemble method. 
The results indicate a strong link between the sim-to-real gap and a decreased model performance, notably exceeding the estimated epistemic uncertainty.

The remainder of this work is organized as follows: Section 2 summarizes the related work, focusing on the concept of predictive uncertainty in more detail. 
Section 3 presents the employed methodology, introducing the examined dataset and experiment setup.
Section 4 presents the gathered results, whereas Section 5 discusses them.
Finally, Section 6 concludes this work by delineating potential limitations and proposing directions for future work.

\section{Related work}

\subsection{Synthetic data generation}
The generation of synthetic training data has emerged as a viable strategy to overcome data scarcity in DL-based applications, particularly in manufacturing domains where the acquisition and annotation of real-world data is costly, time-consuming, or impractical \cite{Moriz:2025,dummel:2021,schieber:2024}. 
Various approaches have been proposed, each with distinct trade-offs regarding realism, flexibility, and applicability to industrial quality control \cite{singh:2025}. 

Generative models, such as Generative Adversarial Networks (GANs), offer a data-driven approach by learning to synthesize images from real data distributions \cite{Niu:2020}. 
While capable of producing visually convincing samples, these methods require a substantial amount of real data for training, which partially undermines the original motivation of synthetic data generation \cite{Man:2022}. 
Furthermore, controlling specific properties of the generated data, such as defect types or precise annotations, remains challenging \cite{singh:2025}, which is especially a drawback for QC applications where precise defect characterization is essential.

Physics-based simulation environments aim to replicate real-world conditions by modeling physical phenomena such as light transport, material properties, and sensor characteristics \cite{Gutierrez:2021}. 
In manufacturing contexts, this can include the simulation of specific MV setups, including camera models and illumination configurations. 
However, these approaches require significant domain expertise to configure and are computationally expensive, making them difficult to adapt across different production environments \cite{Chukwurah:2024}.

CAD-based rendering pipelines leverage existing 3D models of objects and scenes, which are rendered into synthetic images using engines such as Blender or Unity \cite{schieber:2024}. 
This approach is particularly well-suited for industrial QC applications, as CAD models of manufactured parts are typically already available as part of the product development process, eliminating the need for dedicated 3D modeling efforts \cite{dummel:2021}. 
This enables the generation of large-scale, precisely annotated datasets, systematically varying the object's appearance \cite{Schraml:2024}.
However, the realism of the generated data is inherently limited by the fidelity of the rendering process, potentially contributing to a \textit{sim-to-real gap}.

\subsection{Sim-to-real gap}
\label{sec: sim-to-real_gap}
The sim-to-real gap is commonly comprehended as the difference between real and simulated instances, with the real world usually exhibiting a certain behavior that is not covered by the designed simulation, or more broadly, data synthesis approach \cite{akar:2022,schieber:2024,Hinterstoisser:2019}. 
In the context of DL-based image processing, the sim-to-real gap corresponds to the difference between synthetically generated and real images that are used for the training and testing of DL models. 
Due to this difference, synthetically trained models can experience a drop in their predictive performance when evaluated on real datasets \cite{schieber:2024}. 
Although synthetic data samples are generated specifically to address a certain need in the real world, they cannot represent it completely without an extraordinary level of effort with respect to the simulation setup, as demonstrated, e.g., in \cite{akar:2022} for the logistic context. 
From a statistical point of view, the lower performance of synthetically trained DL models for real-world data can be related to the unknown distance between the distributions of the synthetic and real data samples, observed in a high-dimensional feature space \cite{deMelo:2022}. 

To address this distribution difference, \textit{domain adaptation} and \textit{domain randomization} have been proposed as strategies to mitigate the sim-to-real gap \cite{schieber:2024}. 
Domain adaptation methods aim to reduce the distributional shift through the alignment of feature representations. 
While effective, domain adaptation introduces a dependency on real-world target data, diminishing a key advantage of synthetic data generation and adding architectural complexity \cite{Chukwurah:2024}. 
Domain randomization leverages a different strategy by randomizing visual parameters such as textures, lighting, camera poses, and backgrounds during data generation  \hbox{\cite{Moriz:2025,schieber:2024,dummel:2021}}. 
The hypothesis is that sufficient visual diversity during training causes the real-world domain to appear as just another variation within the training distribution, avoiding a sim-to-real gap in the first place. 
Compared to domain adaptation, domain randomization requires no access to real-world data and imposes no constraints on the model architecture, making it particularly useful for industrial applications where real-world data is scarce \cite{Li:2023}. 
However, even with a highly randomized synthetic dataset, the performance drop of synthetically trained models for real-world data can remain \cite{schieber:2024}. 

\subsection{Predictive uncertainty}
The estimation of predictive uncertainty in MV has been extensively studied in safety-critical domains such as autonomous driving, yet its application in QC remains comparatively scarce \cite{araujo:2024,kim:2025}. 
Notably, the few existing works addressing uncertainty in QC settings rely on real-world data, leaving the interplay between synthetic training data and predictive uncertainty unexplored \cite{vanBreugel:2023,Thuy:2024}. 

Predictive uncertainty in DL models is commonly decomposed into two complementary components: \textit{aleatoric uncertainty} and \textit{epistemic uncertainty}. 
Alea-toric uncertainty captures the inherent noise in the data itself, arising from factors such as sensor noise, occlusions, or ambiguous labeling, and cannot be reduced by acquiring more data \cite{Gawlikowski:2022}. 
Epistemic uncertainty reflects the model's lack of knowledge due to limited or unrepresentative training data and can, in principle, be reduced as more informative data becomes available \cite{kendall:2017}.

In the literature, several methods have been proposed to quantify these uncertainties. Bayesian Neural Networks provide a principled probabilistic framework but are computationally expensive and difficult to scale to complex architectures \cite{fakour:2024}. 
Monte Carlo Dropout offers a more practical alternative by performing multiple stochastic forward passes at test time, yet recent work has raised concerns about its reliability, particularly during extrapolation \cite{djupskas:2025}. 
This behavior is especially problematic in the context of models trained with synthetic data, where the sim-to-real gap inherently introduces a distributional shift between training and inference data. 
Ensemble methods, which aggregate predictions from multiple independently trained models, impose no architectural constraints and have been shown to produce well-calibrated uncertainty estimates \cite{fakour:2024}.

It is important to note that ensemble methods primarily quantify epistemic uncertainty, as the variation across ensemble members reflects disagreement arising from limited or unrepresentative training data rather than inherent noise in the data itself \cite{Thuy:2024}.
While this may appear as a limitation, it is well-suited for the investigation pursued in this work. 
When training and testing on synthetic data, the sim-to-real gap does not exist, since the model is exclusively exposed to data from a known and controlled distribution. 
Consequently, any distributional mismatch introduced by the sim-to-real gap is not encoded in the learned representations but rather manifests as a source of epistemic uncertainty during inference on real-world data. 
Therefore, the model's lack of knowledge about real-world characteristics, stemming from never having observed them during training, is the type of uncertainty that ensembles are designed to capture. 
Given these considerations, this work employs the ensemble method for uncertainty estimation, as it combines practical applicability with robust uncertainty quantification while remaining agnostic to the underlying model architecture.

\section{Methodology}
The presence of both the sim-to-real gap and the inherent predictive uncertainty of DL models raises the question of whether an observable performance gap of synthetically trained models could be explained by an increase in the predictive uncertainty caused by a domain shift between the synthetic and real-world data.
To assess this influence, a series of experiments was performed on the public \hbox{T-LESS} dataset following the ensemble method.
The following sections introduce the dataset and the overall experiment setup in more detail.

\subsection{T-LESS dataset}


The T-LESS dataset, originally published by Hodan et al. in \cite{Hodan:2017}, can be found as part of the BOP challenge, which fosters the development of object detection and related tasks in the computer vision domain \cite{Hodan:2020}. 
It comprises images of 30 industry-like objects with no discriminative color or texture and exhibiting high similarity among them \cite{Hodan:2017}. 
As part of the BOP challenge, the T-LESS dataset offers both annotated synthetic and real-world data, where the synthetic data has been generated following a domain randomization strategy.
The dataset provides 50.000 synthetic and 10.080 real-world images, where the real-world images have been acquired through video recordings of 20 scenes of varying complexity \cite{Hodan:2020}.

For this work, the synthetic images are split into three subsets: a training split of 40.000 images, as well as a validation split and a test split of 5.000 images each. 
The additional splitting enables the training of DL models on the training set, the explicit performance monitoring during the training on the validation set, and the performance determination of the final model on the test set, where all sets stem from the same i.i.d. distribution.

\subsection{Experiment setup}
This work focuses on the examination of object detection as one of the most common tasks for industrial MV applications.
While there are plenty of potential models for object detection, the YOLO model family is one of the most widely used, providing good detection results at high processing rates \cite{kim:2025}.
Due to YOLO's end-to-end nature
it is straightforward to train and use, with several open-source libraries such as \textit{Ultralytics} \cite{Jocher:2024} providing further abstraction layers for its application.
Therefore, it is one of the most employed model types for industrial MV applications, as e.g., presented in \cite{Moriz:2025} for defect detection.

In this work, for the four most widely used YOLOv8 model sizes nano (n), small (s), medium (m), and large (l), 25 models were trained on the train-split of the T-LESS dataset, each.
For all experiments, an implementation of the YOLOv8 model from the \textit{Ultralytics} library is used [19], modified only to fit the BOP data structure. 
Each model is trained utilizing the default optimizer and augmentation settings provided by the \textit{Ultralytics} library with a batch size of 8 and an image size of 640x640 px for up to 150 epochs, enabling the early-stopping-mechanism.
The parameters are chosen based on best practice or practical limitations, such as maximum available GPU memory. 
After training, the best model instance (epoch) is chosen based on the highest \textit{mean Average Precision} (mAP) metric score, which is monitored during training on the validation set.
Finally, the models' performance is evaluated on the dedicated test sets, reporting the mAP metric score either averaged over all objects or individually.

\section{Results}
Table \ref{tab:uncertainty_model-wise} shows the mean and standard deviation of the models' performance, calculated using the results of the 25 model instances.
The results show that the overall detection performance increases with model size (complexity) for both dataset types.
Following the ensemble method, the standard deviation can be regarded as the effective predictive uncertainty.
For the synthetic data, the predictive uncertainty is rather low, peaking at 0.28 mAP points for YOLOv8l and being even lower for the other models.
For the real-world data, the predictive uncertainty is with at least 0.41 mAP points notably higher, peaking at 2.78 mAP points for the YOLOv8n model.

\begin{table}[!b]
\centering
\begin{tabular}{c|ccc|ccc}
\toprule
YOLO model & \multicolumn{3}{c}{Model performance (mAP)} & \multicolumn{3}{c}{Standard deviation (mAP)} \\
& synthetic & real-world & $\Delta$ (abs) & synthetic & real-world & $\Delta$ (abs) \\
\midrule
n & 78.83 & 43.45 & 35.38 & 0.06 & 2.78 & 2.72 \\
s & 81.77 & 71.45 & 10.32 & 0.07 & 0.71 & 0.64 \\
m & 83.87 & 78.83 & 5.04 & 0.05 & 0.30 & 0.25 \\
l & 85.64 & 81.28 & 4.36 & 0.28 & 0.41 & 0.13 \\
\bottomrule
\end{tabular}
\caption{Mean and standard deviation values, aggregated over the results of 25 model instances per YOLO model type. Following the ensemble method, the standard deviation can be interpreted as the effective predictive uncertainty.}
\label{tab:uncertainty_model-wise}
\end{table}

The results shown in Table \ref{tab:uncertainty_model-wise} are subject to a double averaging: 
First, following the definition of mAP, the single-class detection scores are averaged to yield an overall value for the dataset. 
Second, following the ensemble method, the results of the individual models are averaged over the number of models (ensemble members) to gain insights regarding the uncertainty. 
Consequently, this double averaging could mask the influence of particular object classes and/or individual models. 
Therefore, the analysis in this work is expanded by examining the distribution of the \textit{inter-class} and \textit{intra-class} variance of the model results.

\begin{figure}[!b]
    \centering
\begin{subfigure}{0.48\textwidth}
    \centering
    \includegraphics[width=\linewidth, trim={0 0 0 0.55cm}, clip]{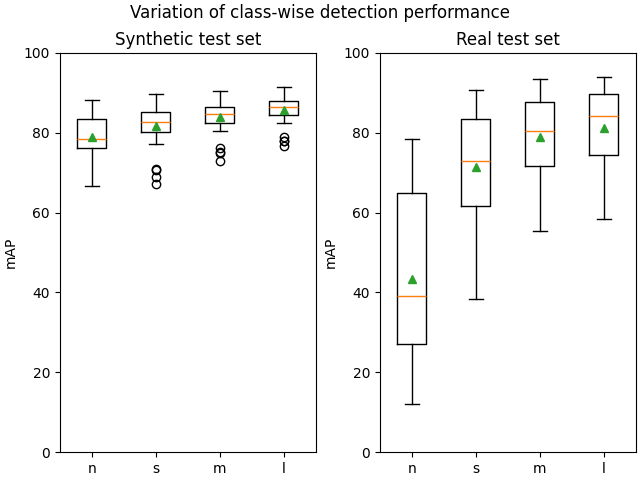}
    \caption{
    Distribution of class-wise mean detection scores for both datasets, illustrating the inter-class performance variance.
    }
\end{subfigure}
\hfill
\begin{subfigure}{0.48\textwidth}
    \centering
    \includegraphics[width=\linewidth, trim={0 0 0 0.55cm}, clip]{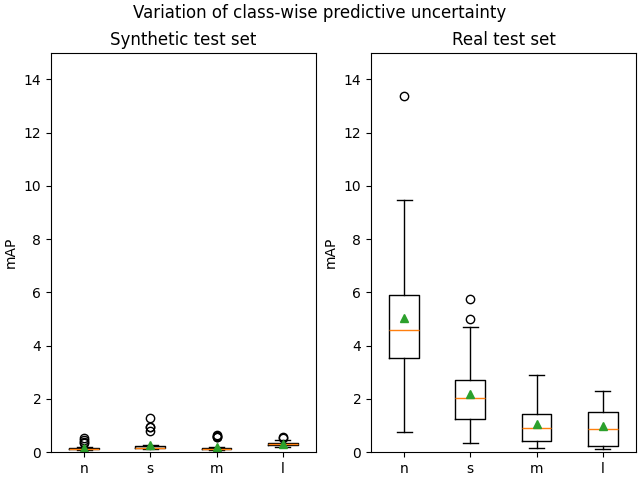}
    \caption{
    Distribution of class-wise standard deviation values for both datasets, illustrating the intra-class performance variance.
    }
\end{subfigure}
\caption{
Analysis of the individual class detection performance of the four evaluated YOLO models. The boxplots show the distributions of the mean and standard deviation values of class-wise mAP scores for both the synthetic and the real-world datasets, aggregated using the 25 ensemble members. 
}
    \label{fig:uncertainty_class-wise}
\end{figure}

The inter-class variance distribution is determined by calculating the mean detection score for each class separately over the ensemble, aggregating thus 30 class-wise mean detection results per YOLO model type. 
The distribution is shown in Fig. \ref{fig:uncertainty_class-wise} (a) in the form of boxplots. 
The mean of these aggregated values (cf. green triangles) corresponds to the mean performance scores introduced in Table \hbox{\ref{tab:uncertainty_model-wise}}. 
The distribution in \hbox{Fig. \ref{fig:uncertainty_class-wise} (a)} illustrates the capability of the model architectures to detect all object classes equally well. 
Larger distributions and/or outliers could indicate either difficulties of the model type to learn individual classes or the imbalance of these classes in the used training or test set.
In general, \hbox{Fig. \ref{fig:uncertainty_class-wise} (a)} shows the same dynamic observed in \hbox{Table \ref{tab:uncertainty_model-wise}}, with the average model performance increasing with increasing model complexity.
Yet, the variation of detection scores for the individual classes notably exceeds the observed level of predictive uncertainty even for the best model (YOLOv8l) on the synthetic test set.
For the real-world data, the class detection scores vary even more strongly. 

The intra-class variance is determined by observing the variance of the class-wise detection results over the ensemble, yielding thus also 30 class-wise scores per YOLO model type.
The intra-class variance illustrates the model architecture's robustness in learning a particular class.
A narrow distribution indicates that the model ends up in a similar local optimum for the considered use case, regardless of the initial state.
On the other hand, larger intra-class variance values could indicate inherent problems in the representation of classes within the training or test dataset, or an instability in detection due to different local optima of the model.
The distribution of intra-class values is shown in \hbox{Fig. \ref{fig:uncertainty_class-wise} (b)}.   
Although being slightly higher than the predictive uncertainty, the intra-class variance on the synthetic set is considered negligible, being notably lower than the observed sim-to-real performance gap.
For the real-world data, the improvement trend from \hbox{Fig. \ref{fig:uncertainty_class-wise} (a)} is observable as well, showing a decreasing variance with increasing model complexity.
However, similar to Fig. \ref{fig:uncertainty_class-wise} (a), the similarity in the distributions of the YOLOv8m and YOLOv8l results indicates the arrival at a certain performance plateau, which is most likely linked to a certain data characteristic. 


\section{Discussion}
The results in Table \ref{tab:uncertainty_model-wise} show a notable difference between the models' performance on synthetic and on real-world data. 
In theory, this performance gap can be due to various reasons, such as a present class imbalance, the characteristics of the real-world dataset, or the sim-to-real gap.

The results for the synthetic test set, which were created from the same distribution as the training set, exhibit very low values for the ensemble-based uncertainty estimation. 
Since ensembles primarily capture epistemic uncertainty, it can be concluded that the epistemic contribution to a model's individual performance can almost be neglected as the models converge consistently despite different random initializations. 
This confirms a high robustness of the training process and suggests that the synthetic training distribution provides sufficient information for the models to learn the target relationships. 
At the same time, the models' evaluation on real-world data shows an increase in the predictive uncertainty compared to the results on the synthetic test set.
Since no other factors were changed, this increase can most likely be attributed to the domain shift between the synthetic and real-world data.
By implication, this means that the employed domain randomization strategy is not capable of increasing the feature space to a sufficient degree to cover the real-world data completely.
Hence, the increase in predictive uncertainty confirms the presence of a distributional shift caused by the sim-to-real gap between the two examined datasets. 
Yet, the increase in predictive uncertainty alone does not explain the observable performance gap, with the latter being several orders of magnitude larger than the former.
Since all ensemble members are trained on the same synthetic data distribution, they converge towards similar feature representations that are consistently worse for the real-world domain. 
Consequently, the models exhibit a systematic shared error that does not produce disagreement among the ensemble members. 
The remaining performance gap is therefore attributed to a systematic distributional bias, which is linked to the sim-to-real gap.

To gain further insights, the detailed analysis of class-wise results displayed in Fig. \ref{fig:uncertainty_class-wise} was performed.
The low values of the intra-class performance variance on the synthetic test set confirm the low epistemic uncertainty, demonstrating a high robustness of the models' learning process against randomization in the initialization. 
However, the distribution of the inter-class variance values shows much larger performance deviations even for the synthetic set.
This indicates an inherent data issue related to the object classes, such as a disproportional occurrence of classes due to their size or a high inter-class similarity leading to mutual misclassifications. 
The evaluations on the real-world dataset show an even stronger influence of individual classes.
Here, a clear distinction between the influence of the sim-to-real gap and other data-related factors, such as class-imbalance, cannot be made.
Yet, the similarity of the analyzed distributions for the YOLOv8m and the YOLOv8l models suggests the convergence towards a certain plateau, which in turn indicates a dataset-specific issue that cannot be surpassed without explicit consideration. 


\section{Conclusion and Future Work}
The presented work examines the link between predictive uncertainty and the sim-to-real gap that occurs for DL models trained on synthetic data. 
The results indicate that while the sim-to-real gap most likely affects the model performance, the influence attributed to model-related uncertainty factors is comparatively negligible.
Despite being unable to quantify the influence of the sim-to-real gap on the performance completely, the presented approach still constitutes a valuable asset: 
Since the determination of the predictive uncertainty through the ensemble method does not rely on the actual performance values, it is not influenced by the sim-to-real gap.
Yet, its increase compared to the synthetic set is linked to the present domain shift, which in turn is caused by the sim-to-real gap.
Hence, the predictive uncertainty can serve as an instrument for an assessment of the domain shift, and, indirectly, also of the sim-to-real gap. 

Future work will focus on a detailed analysis of the data-related factors influencing the performance, stemming both from the synthetic data generation approach and from implicit, use-case-specific characteristics.


%

\bibliographystyle{styles/bibtex/spmpsci_unsrt}
\bibliography{references}

@Article{araujo:2024,
AUTHOR = {Araújo, Bernardo and Teixeira, João F. and Fonseca, Joaquim and Cerqueira, Ricardo and Beco, Sofia C.},
TITLE = {The Road to Safety: A Review of Uncertainty and Applications to Autonomous Driving Perception},
JOURNAL = {Entropy},
VOLUME = {26},
YEAR = {2024},

}

@article{kim:2025,
author = {Seongje Kim and Thong Phi Nguyen and Jonghun Yoon},
title = {A Systematic Review of Machine Vision Applications in Factory and Manufacturing Processes: From Quality Control to Predictive Diagnostics},
journal = {International Journal of Precision Engineering and Manufacturing-Green Technology},
year = {2025},
}

@misc{vanBreugel:2023,
      title={Beyond Privacy: Navigating the Opportunities and Challenges of Synthetic Data}, 
      author={Boris van Breugel and Mihaela van der Schaar},
      year={2023},
      journal={arXiv}
}

@article{singh:2025,
title = {A comprehensive survey of image synthesis approaches for Deep Learning-based surface defect detection in manufacturing},
journal = {Computers in Industry},
year = {2025},
author = {Aru Ranjan Singh and Sumit Hazra and Abhishek Goswami and Kurt Debattista and Thomas Bashford-Rogers},
}

@ARTICLE{Niu:2020,
  author={Niu, Shuanlong and Li, Bin and Wang, Xinggang and Lin, Hui},
  journal={IEEE Transactions on Automation Science and Engineering}, 
  title={Defect Image Sample Generation With GAN for Improving Defect Recognition}, 
  year={2020},
  pages={1611-1622},}

@article{Gutierrez:2021,
  author       = {Pierre Gutierrez and
                  Maria Luschkova and
                  Antoine Cordier and
                  Mustafa Shukor and
                  Mona Schappert and
                  Tim Dahmen},
  title        = {Synthetic training data generation for deep learning based quality
                  inspection},
  journal      = {arXiv},
  year         = {2021},

}

@misc{kendall:2017,
      title={What Uncertainties Do We Need in Bayesian Deep Learning for Computer Vision?}, 
      author={Alex Kendall and Yarin Gal},
      year={2017},
      journal={arXiv},
}

@article{Thuy:2024,
   title={Fast and reliable uncertainty quantification with neural network ensembles for industrial image classification},
   journal={Annals of Operations Research},
   publisher={Springer Science and Business Media LLC},
   author={Thuy, Arthur and Benoit, Dries F.},
   year={2024},}

@misc{djupskas:2025,
      title={Unreliable Uncertainty Estimates with Monte Carlo Dropout}, 
      author={Aslak Djupskås and Alexander Johannes Stasik and Signe Riemer-Sørensen},
      year={2025},
      archivePrefix={arXiv},
}

@Article{Schraml:2024,
AUTHOR = {Schraml, Dominik and Notni, Gunther},
TITLE = {Synthetic Training Data in AI-Driven Quality Inspection: The Significance of Camera, Lighting, and Noise Parameters},
JOURNAL = {Sensors},
YEAR = {2024},
}

@article{Li:2023,
title = {A domain adaptation YOLOv5 model for industrial defect inspection},
journal = {Measurement},
year = {2023},
author = {Chen Li and Haoxin Yan and Xiang Qian and Shidong Zhu and Peiyuang Zhu and Chengwei Liao and Haoyang Tian and Xiu Li and Xiaohao Wang and Xinghui Li},
}

@misc{fakour:2024,
      title={A Structured Review of Literature on Uncertainty in Machine Learning \& Deep Learning}, 
      author={Fahimeh Fakour and Ali Mosleh and Ramin Ramezani},
      year={2024},
      journal={arXiv}
}

@Article{Man:2022,
AUTHOR = {Man, Keith and Chahl, Javaan},
TITLE = {A Review of Synthetic Image Data and Its Use in Computer Vision},
JOURNAL = {Journal of Imaging},
YEAR = {2022},
}

@article{Ahangar:2025,
author = {M. Nadeem Ahangar and Z. A. Farhat and Aparajithan Sivanathan},
title = {AI Trustworthiness in Manufacturing: Challenges, Toolkits, and the Path to Industry 5.0},
journal = {Sensors (Basel, Switzerland)},
year = {2025},
}

@INPROCEEDINGS{akar:2022,
  author={Akar, Chafic Abou and Tekli, Jimmy and Jess, Daniel and Khoury, Mario and Kamradt, Marc and Guthe, Michael},
  booktitle={2022 35th SIBGRAPI Conference on Graphics, Patterns and Images (SIBGRAPI)}, 
  title={Synthetic Object Recognition Dataset for Industries}, 
  year={2022},
  pages={150-155},}

@article{schieber:2024,
title = {Indoor Synthetic Data Generation: A Systematic Review},
journal = {Computer Vision and Image Understanding},
year = {2024},
author = {Hannah Schieber and Kubilay Can Demir and Constantin Kleinbeck and Seung Hee Yang and Daniel Roth},
}

@INPROCEEDINGS {Hinterstoisser:2019,
author = { Hinterstoisser, Stefan and Pauly, Olivier and Heibel, Hauke and Martina, Marek and Bokeloh, Martin },
booktitle = { 2019 IEEE/CVF International Conference on Computer Vision Workshop (ICCVW) },
title = {{ An Annotation Saved is an Annotation Earned: Using Fully Synthetic Training for Object Detection }},
year = {2019},
pages = {2787-2796},}

@InProceedings{dummel:2021,
author="D{\"u}mmel, Johannes
and Kostik, Valentin
and Oellerich, Jan",
title="Generating Synthetic Training Data for Assembly Processes",
booktitle="Advances in Production Management Systems. Artificial Intelligence for Sustainable and Resilient Production Systems",
year="2021",
publisher="Springer International Publishing",
pages="119-128",
}

@article{deMelo:2022,
title = {Next-generation deep learning based on simulators and synthetic data},
journal = {Trends in Cognitive Sciences},
pages = {174-187},
year = {2022},
author = {Celso M. {de Melo} and Antonio Torralba and Leonidas Guibas and James DiCarlo and Rama Chellappa and Jessica Hodgins},
}

@article{Chukwurah:2024,
  title={Sim-to-real transfer in robotics: Addressing the gap between simulation and real-world performance},
  author={Chukwurah, Naomi and Adebayo, Abiodun Sunday and Ajayi, Olanrewaju Oluwaseun},
  journal={International Journal of Robotics and Simulation},
  pages={89-102},
  year={2024}
}

@article{Baltag:2022,
author = {Baltag, Ciprian-Daniel and Morariu, Cristin Olimpiu},
title = {Quality and Risk Management in Industrial Production Systems: A Literature Review},
journal = {Bulletin of the Polytechnic Institute of Iași. Machine constructions Section},
year = {2022},
pages = {75-112}
}

@misc{Gawlikowski:2022,
      title={A Survey of Uncertainty in Deep Neural Networks}, 
      author={Jakob Gawlikowski and Cedrique Rovile Njieutcheu Tassi and Mohsin Ali and Jongseok Lee and Matthias Humt and Jianxiang Feng and Anna Kruspe and Rudolph Triebel and Peter Jung and Ribana Roscher and Muhammad Shahzad and Wen Yang and Richard Bamler and Xiao Xiang Zhu},
      year={2022},
      journal={arXiv},
}

@misc{Hodan:2017,
 author = {Hodan, Tomas and Haluza, Pavel and Obdrzalek, Stepan and Matas, Jiri and Lourakis, Manolis and Zabulis, Xenophon},
 date = {2017},
 title = {T-LESS: An RGB-D Dataset for 6D Pose Estimation of Texture-less Objects},
 journal={arXiv},
}

@misc{Hodan:2020,
 author = {Hodan, Tomas and Sundermeyer, Martin and Drost, Bertram and Labbe, Yann and Brachmann, Eric and Michel, Frank and Rother, Carsten and Matas, Jiri},
 year = {2020},
 title = {BOP Challenge 2020 on 6D Object Localization},
 journal = {arXiv},
}

@misc{Jocher:2024,
 author = {{Glenn Jocher} and {Jing Qiu}},
 year = {2024},
 title = {Ultralytics YOLOv8},
 url = {https://github.com/ultralytics/ultralytics}
}

@INPROCEEDINGS{Moriz:2025,
  author={Moriz, Alexander and Wolfschläger, Dominik and Svetlakov, Andrei and Suchodolak, Miguel and Byl, Achim and Wolf-Monheim, Friedrich and Wege, Kai and Carey, Eilis and Schmitt, Robert H.},
  booktitle={2025 IEEE 30th International Conference on Emerging Technologies and Factory Automation (ETFA)}, 
  title={Generation of Synthetic Data for DL-based Defect Detection in the Automotive Context}, 
  year={2025},
  volume={},
  number={},
  pages={1-8},
 }

\end{document}